\documentclass[10pt,twocolumn,letterpaper]{article}

\usepackage[pagenumbers]{iccv} 

\usepackage{multirow}
\usepackage{amssymb}
\usepackage{graphicx}
\usepackage{utfsym}

\definecolor{iccvblue}{rgb}{0.21,0.49,0.74}
\usepackage[pagebackref,breaklinks,colorlinks,allcolors=iccvblue]{hyperref}

\def\paperID{11007} 
\def\confName{ICCV}
\def\confYear{2025}

\title{MMAT-1M: A Large Reasoning Dataset for Multimodal Agent Tuning}

\author{
Tianhong Gao\thanks{Equal contribution.} \quad
Yannian Fu\footnotemark[1]\hspace{0.4em}\thanks{Corresponding author.} \quad
Weiqun Wu \quad
Haixiao Yue \quad
Shanshan Liu \quad
Gang Zhang \\
Baidu Inc. \\
}

\begin{document}
\maketitle

\begin{abstract}
Large Language Models (LLMs), enhanced through agent tuning, have demonstrated remarkable capabilities in Chain-of-Thought (CoT) and tool utilization, significantly surpassing the performance of standalone models. However, the multimodal domain still lacks a large-scale, high-quality agent tuning dataset to unlock the full potential of multimodal large language models. To bridge this gap, we introduce MMAT-1M, the first million-scale multimodal agent tuning dataset designed to support CoT, reflection, and dynamic tool usage. Our dataset is constructed through a novel four-stage data engine: 1) We first curate publicly available multimodal datasets containing question-answer pairs; 2) Then, leveraging GPT-4o, we generate rationales for the original question-answer pairs and dynamically integrate API calls and Retrieval Augmented Generation (RAG) information through a multi-turn paradigm; 3) Furthermore, we refine the rationales through reflection to ensure logical consistency and accuracy, creating a multi-turn dialogue dataset with both Rationale and Reflection (RR); 4) Finally, to enhance efficiency, we optionally compress multi-turn dialogues into a One-turn Rationale and Reflection (ORR) format. By fine-tuning open-source multimodal models on the MMAT-1M, we observe significant performance gains. For instance, the InternVL2.5-8B-RR model achieves an average improvement of 2.7\% across eight public benchmarks and 8.8\% on the RAG benchmark Dyn-VQA, demonstrating the dataset's effectiveness in enhancing multimodal reasoning and tool-based capabilities. 
The dataset is publicly available at \mbox{\href{https://github.com/VIS-MPU-Agent/MMAT-1M}{https://github.com/VIS-MPU-Agent/MMAT-1M}}.
\end{abstract}

\begin{figure}[t]
    \centering
    \includegraphics[width=\linewidth]{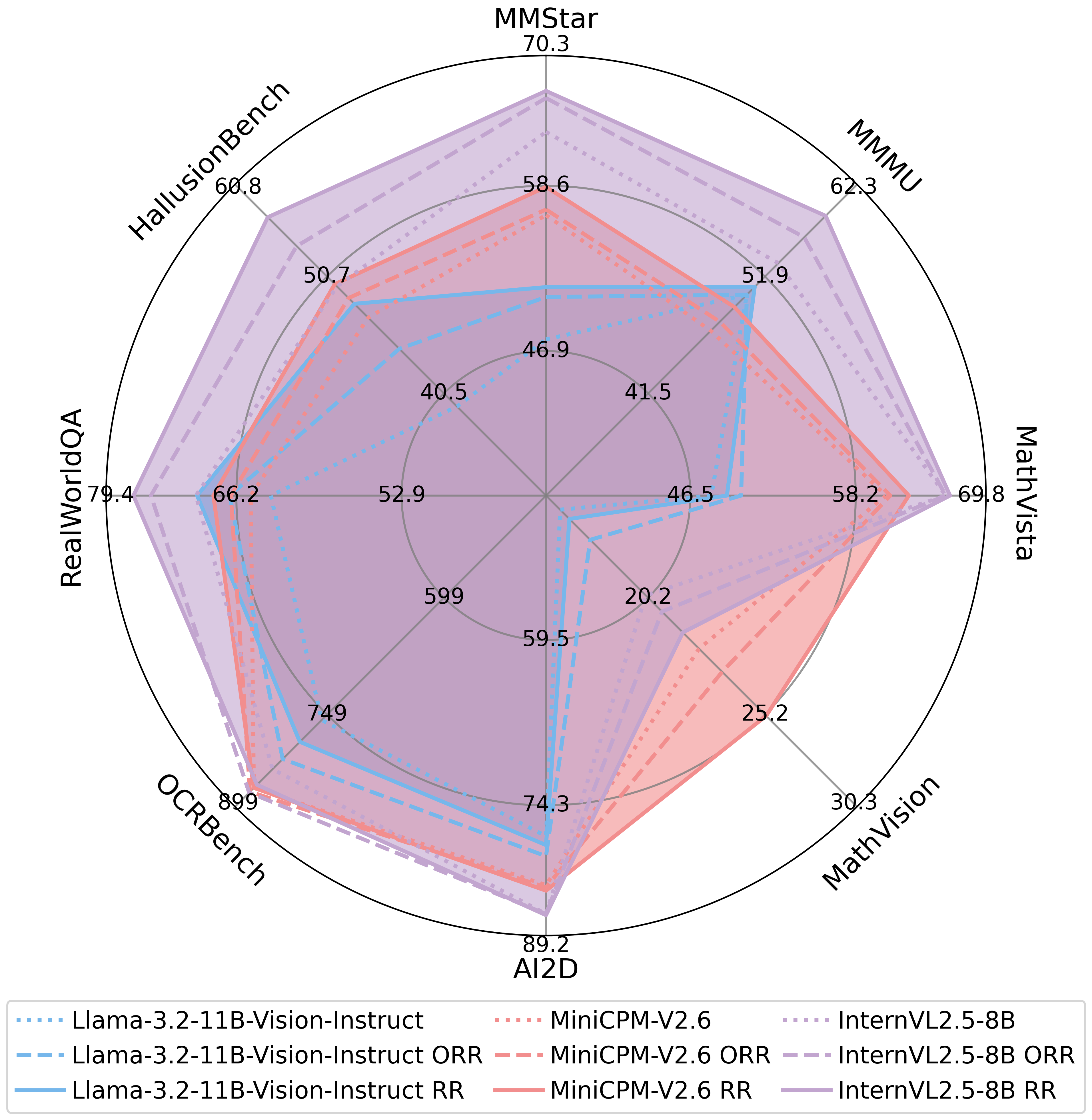}
    \caption{Performance comparison of multimodal large language models fine-tuned on MMAT-1M dataset using One-turn Rationale and Reflection (ORR) and Rationale and Reflection (RR) across eight benchmarks. Both strategies significantly boost performance,  demonstrating the effectiveness of structured reasoning and MMAT-1M.}
    \label{fig:radar}
\end{figure}

\section{Introduction}
In recent years, Multimodal Large Language Models (MLLMs) exemplified by GPT-4o~\cite{shahriar2024putting}, Gemini~\cite{team2023gemini}, the QwenVL series~\cite{bai2023qwen,Qwen2-VL,Qwen2.5-VL}, the InternVL series~\cite{chen2024internvl,chen2024far,chen2024expanding}, and the LLaVA series~\cite{liu2023visual,liu2024improved} have made remarkable strides. To further enhance the reasoning and problem-solving capabilities of these models, integrating Chain-of-Thought (CoT) reasoning and external tools has proven to be an effective approach, commonly referred to as ``Agents''. Agents operate through two primary methods: instruction-driven~\cite{shen2023hugginggpt,yang2023gpt4tools,wu2023visual,yao2023react,gao2023assistgpt} and tuning-driven~\cite{zeng2023agenttuning,chen2023fireact,chen2024agent,yin2024agent,wang2024llms}. The former involves designing prompts to enable LLMs to plan, reason, and utilize tools, which demands strong prompt comprehension. The latter employs specialized datasets to fine-tune models, empowering even smaller models to achieve agent capabilities comparable to proprietary large models. Consequently, agent tuning has emerged as a prominent and promising research direction.

In terms of existing research, several representative works have emerged in the field of multimodal agent tuning. For instance, LLaVA-Plus~\cite{liu2024llava} converts LLaVA-158K dataset into a tool-use instruction format with 117K samples through both user-oriented and skill-oriented dialogues, and T3-Agent~\cite{gao2024multi} constructs the MM-Traj dataset that contains 20K multimodal tasks with tool-usage trajectories. However, existing datasets commonly suffer from three critical shortcomings: (1) They exhibit a relatively homogeneous distribution, limiting improvements to diverse benchmarks; (2) They lack mechanisms for reflecting on errors induced by visual tools, resulting in weak model robustness against interference; (3) They are deficient in flexible reasoning and tool-usage mechanisms, reducing their feasibility for real-world applications. Consequently, building a large-scale tuning dataset that effectively addresses these challenges—diversity, robustness, and flexibility—has emerged as a critical breakthrough for advancing the field.

To overcome these bottlenecks, we propose Multi-Modal Agent Tuning—One Million (MMAT-1M), which, to the best of our knowledge, is the first million-scale multimodal agent tuning dataset including diverse fundamental visual tasks. Building on publicly available multimodal datasets, we design a four-stage data synthesis framework. First, we compile publicly accessible multimodal datasets that encompass question-answer pairs. To ensure consistency in input and output formats across diverse multimodal datasets, we adapt the prompts for both inputs and outputs. Then we generate iterative trajectories using CoT reasoning and dynamic API calls, incorporating functionalities such as Image Caption, Optical Character Recognition (OCR), Open-Vocabulary Object Detection (OVD), Face Detection, and RAG. Next, we evaluate these trajectories for logical inconsistencies and refine those requiring modification through a reflection process. To enhance practical flexibility, we optionally consolidate iterative trajectories into a one-turn format and prepend tool-usage results to the input. Experimental results demonstrate that models fine-tuned with the MMAT-1M dataset exhibit significant performance advantages. As Figure~\ref{fig:radar} shows, after training on our two formats of datasets, all three mainstream open-source models achieve better performance compared to the baseline. Taking the InternVL2.5-8B-RR model as an example, it achieves an average improvement of 2.7\% across eight publicly available multimodal benchmarks compared to the baseline model. Furthermore, on the Dyn-VQA benchmark, which requires multi-hop reasoning and web search capabilities, it demonstrates an improvement of 8.8\%.

The main contributions of this study can be summarized as follows:
(1) We propose the first million-scale multimodal agent tuning dataset, MMAT-1M, addressing a critical gap in the domain of multimodal agent tuning.
(2) We establish a reflection mechanism that effectively mitigates logical errors in the reasoning process, significantly enhancing the model's robustness.
(3) We offer datasets in both one-turn and iterative formats, providing flexibility to balance precision and efficiency in practical applications.

\section{Related Work}

\noindent\textbf{LLM-based Agents} LLM-based agents are primarily large models that harness the instruction-following capabilities of LLMs to develop advanced reasoning and tool-usage functionalities. Notable frameworks in this domain include HuggingGPT~\cite{shen2023hugginggpt}, GPT4Tools~\cite{yang2023gpt4tools}, VisualChatGPT~\cite{wu2023visual}, among others. ReAct~\cite{yao2023react}, for instance, introduces a general paradigm that integrates CoT reasoning with action execution to address a broad spectrum of reasoning and decision-making challenges. Similarly, AssistGPT~\cite{gao2023assistgpt}, proposes a ``Learner'' module that analyzes the prediction process and facilitates reflection, aligning with ReAct's methodology. However, these approaches heavily rely on the instruction comprehension capabilities of LLMs, which restricts their effectiveness in handling longer or more complex reasoning tasks. Additionally, the high computational costs associated with invoking large models further raise the barrier to practical application.

\noindent\textbf{Multimodal Agent Tuning.} Agent tuning is a specialized subfield of language model fine-tuning, focused on enhancing the capabilities of LLMs in areas such as planning, reasoning, and tool usage. Among the earliest works in this domain are AgentTuning~\cite{zeng2023agenttuning} and Fireact~\cite{chen2023fireact}, which laid the foundation for subsequent advancements in agent tuning. Subsequently, many efforts are dedicated to advancing agent tuning~\cite{chen2024agent,yin2024agent,wang2024llms,song2024trial}. However, these methods primarily concentrate on optimizing LLMs, which, when applied in the multimodal domain, can only access information through multimodal tools. To address this limitation, several studies have explored multimodal agent tuning to improve reasoning and tool usage for multimodal challenges. For instance, LLaVA-Plus~\cite{liu2024llava} represents the first attempt to train a multimodal assistant through visual instruction tuning, enabling it to learn tool usage effectively. Similarly, MLLM-Tool~\cite{wang2024mllm} is an agent system that integrates multimodal encoders with open-source LLMs to perceive and process instructions based on visual or audio inputs. Additionally, T3-Agent~\cite{gao2024multi} generates a diverse range of multimodal tasks with detailed trajectories and leverages this data to fine-tune Vision-Language Models (VLMs) for enhanced tool utilization.

\noindent\textbf{Multimodal Agent and CoT Dataset.} To achieve strong performance in multimodal agent tuning, several datasets have been developed to optimize agents using diverse approaches. For instance, LLaVA-Plus transforms the LLaVA-158K dataset into a tool-use instruction format. Similarly, MLLM-Tool curates instruction-answer pairs encompassing 29 tasks sourced from HuggingFace. Meanwhile, T3-Agent introduces MM-Traj, a dataset comprising 20K trajectories, generated through a novel data collection pipeline. Moreover, some agents, such as OmniSearch~\cite{li2024benchmarking}, have designed the Dyn-VQA benchmark to evaluate capabilities in RAG and multi-hop reasoning tasks.
In addition to these multimodal agent datasets, several multimodal CoT datasets share similar construction methodologies but lack explicit information on tool usage, such as LLaVA-CoT~\cite{xu2024llava}, Visual-CoT~\cite{shao2024visual}, and $\text{M}^{3}\text{CoT}$~\cite{chen2024m}.

\section{MMAT-1M Dataset}
In this section, we provide a comprehensive introduction to MMAT-1M, detailing its key components and methodologies. The discussion is structured into three parts: (1) an overview of the dataset, which outlines its scope, composition, and significance (Section~\ref{sec:overview}); (2) the data engine, which describes the iterative framework for generating and refining high-quality trajectories (Section~\ref{sec:data_engine}); and (3) the multimodal agent tuning method, which explains the approaches for enhancing reasoning and tool-usage capabilities (Section~\ref{sec:mmat}).

\begin{figure*}[htbp]
    \centering
    \includegraphics[width=0.9\textwidth]{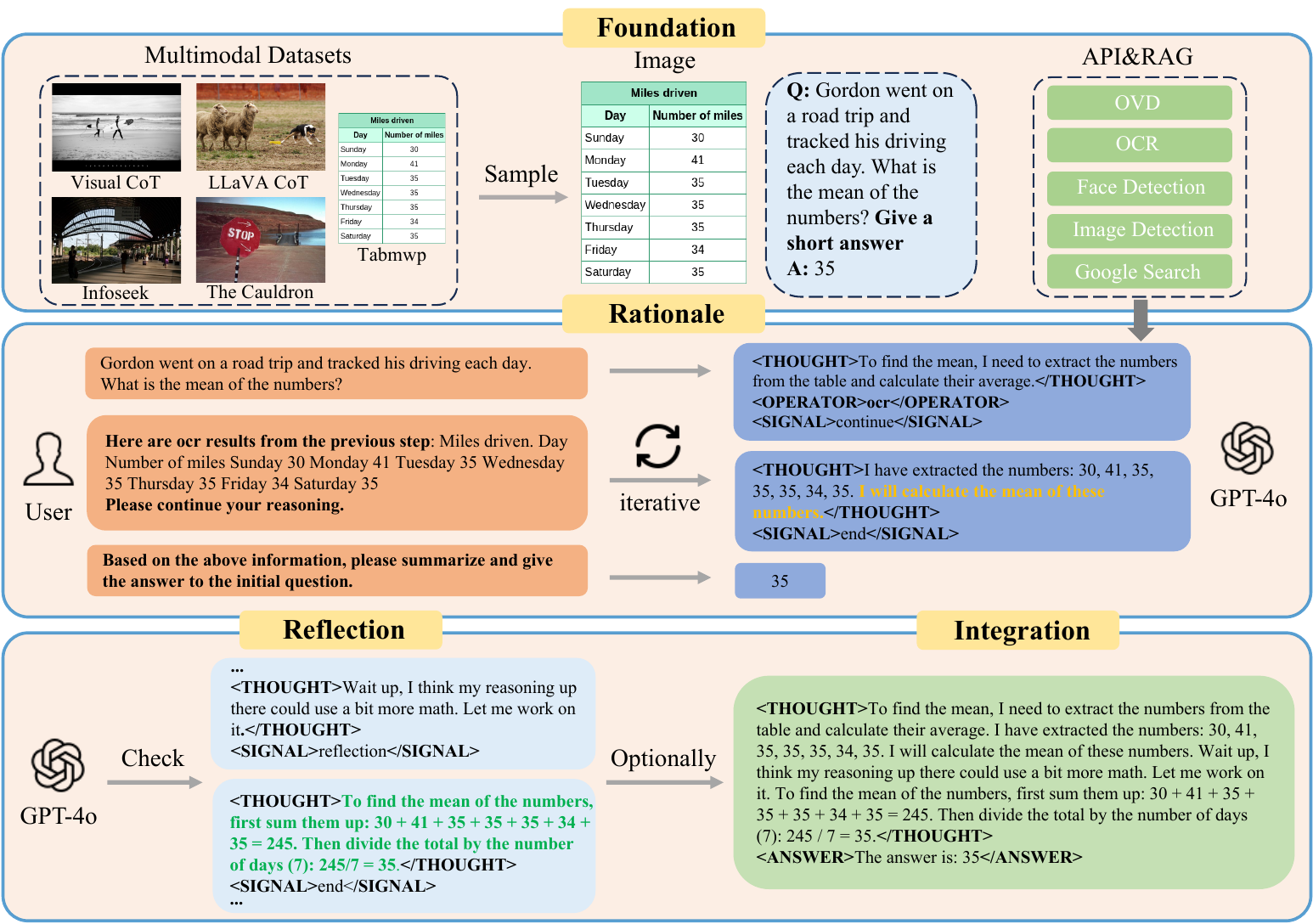}
    \caption{The data engine pipeline follows four stages: foundation, rationale generation, reflection, and trajectory integration. It generates datasets in two formats (RR and ORR) achieving a balance between precision and efficiency.}
    \label{fig:data_engine_figure}
\end{figure*}

\subsection{Overview of MMAT-1M}\label{sec:overview}
\begin{table}[t]
  \footnotesize
  \centering
  \begin{tabular}{llc}
    \toprule
    Statistics & Component & Number \\
    \midrule
    \multirow{6}{*}{Dataset Composition} 
    & Visual CoT~\cite{shao2024visual} & 434265 \\
    & LLaVA-CoT~\cite{xu2024llavacot} & 98561 \\
    & The Cauldron~\cite{laurenccon2024matters} & 215680 \\
    & TabMWP~\cite{lu2022dynamic} & 23059 \\
    & Infoseek~\cite{chen2023can} & 131400 \\
    \midrule
    \multirow{3}{*}{Dialogue Turns} 
    & 1 turn & 846389 \\
    & 2 turns & 28646 \\
    & 3+ turns & 27930 \\
    \midrule
    \multirow{3}{*}{Rationale Steps} 
    & 2 turn & 7909 \\
    & 3 turns & 763212 \\
    & 4 turns & 221440 \\
    & 5+ turns & 97702 \\
    \midrule
    \multirow{5}{*}{Operator Calls} 
    & Image Caption & 620644 \\
    & OVD & 156237 \\
    & OCR & 471866 \\
    & Face Detection & 20077 \\
    & RAG & 205682 \\
    \midrule
    \multirow{2}{*}{Reflection Calls} 
    & General & 46508 \\
    & Math & 11139 \\
    \bottomrule
  \end{tabular}
  \caption{Key statistics of the MMAT-1M dataset.}
  \label{tab:dataset_analysis}
\end{table}
To build a diverse and comprehensive MMAT-1M dataset, we consolidate data from five distinct sources. These sources encompass a wide range of critical domains in multimodal tasks, including visual understanding, logical reasoning, mathematical computation, and knowledge retrieval. This integration ensures both the diversity and completeness of the dataset. The details of each dataset are as follows:

\textbf{Visual CoT}~\cite{shao2024visual} encompasses a variety of tasks, such as document parsing, fine-grained understanding, general visual question answering (VQA), chart analysis, and relational reasoning. Its primary objective is to strengthen models' capabilities in focusing on localized visual regions and executing step-by-step reasoning processes.
\begin{table}[t!]
\centering
\footnotesize
\resizebox{\linewidth}{!}{
\begin{tabular}{lcccccc}
    \toprule
    Dataset & Size & APIs & Online Search & CoT & Reflection & Turns\\
    \midrule
    LLaVA-Plus-v1~\cite{liu2024llava} & 117K & \checkmark & \scalebox{0.75}{\usym{2613}} & \checkmark & \scalebox{0.75}{\usym{2613}} & multiple \\
    Visual CoT~\cite{shao2024visual} & 438K & \scalebox{0.75}{\usym{2613}} & \scalebox{0.75}{\usym{2613}} & \checkmark & \scalebox{0.75}{\usym{2613}} & one \\
    LLaVA-CoT~\cite{xu2024llavacot} & 100K & \scalebox{0.75}{\usym{2613}} & \scalebox{0.75}{\usym{2613}} & \checkmark & \scalebox{0.75}{\usym{2613}} & one \\
    MM-Traj~\cite{gao2024multi} & 20K & \checkmark & \checkmark & \checkmark & \scalebox{0.75}{\usym{2613}} & one \\
    \midrule
    MMAT-1M & 1M & \checkmark & \checkmark & \checkmark & \checkmark & one\&multiple \\
    \bottomrule
\end{tabular}
}
\caption{Comparison of MMAT-1M with other training datasets.}
\label{tab:dataset_comparison}
\end{table}
\textbf{LLaVA-CoT}~\cite{xu2024llavacot} places a strong emphasis on complex reasoning and systematic thinking. It tackles a range of tasks, including general VQA, scientific reasoning, mathematical reasoning, and document understanding, aiming to enhance models' hierarchical reasoning capabilities and improve their interpretability.
\textbf{The Cauldron}~\cite{laurenccon2024matters} incorporates a wide array of multimodal data types, including interleaved text-image documents, text-image pairs, OCR-processed documents, and tables or charts. The diversity of its data sources and task designs plays a pivotal role in advancing models' generalization capabilities, particularly in the integration of visual and linguistic information.
~\textbf{TabMWP}~\cite{lu2022dynamic} focuses on mathematical reasoning tasks that integrate both textual and tabular data, seeking to improve models' table parsing, numerical computation, and complex reasoning capabilities.
\textbf{Infoseek}~\cite{chen2023can} is centered on visual information-seeking question answering, designed to assess and enhance the performance of multimodal models in knowledge-intensive visual question-answer tasks. These tasks demand fine-grained reasoning that extends beyond common sense and often relies on external knowledge bases for accurate responses.

The statistical information of the MMAT-1M dataset is shown in Table~\ref{tab:dataset_analysis}. The dataset comprises a total of 1,090,263 question-answer pairs and 902,965 dialogues, distributed across distinct subsets to ensure diversity in data sources. The second row of the table shows the number of dialogue turns in the original data, which shows that the one-turn dialogues represent the majority of samples, while the multi-turn dialogues are comparatively less frequent.
In terms of reasoning complexity, the majority of data samples involve two-step and three-step reasoning processes, which serve as the foundational level of reasoning. In contrast, tasks requiring more intricate, multi-step reasoning constitute a smaller proportion, highlighting the dataset's inclusion of both basic and advanced cognitive challenges. Meanwhile, among a wide range of operator calls, the invocation of Image Caption and OCR is relatively high, indicating the demand for basic information of images and text in the reasoning process. RAG and OVD also account for a notable proportion of operator invocations. Furthermore, the reflection section encompasses both general reflection and mathematical reasoning reflection, comprising a total of approximately 57k data points. In summary, MMAT-1M is distinguished by its large-scale data volume, diverse task coverage, and hierarchical reasoning depth, collectively establishing a robust and flexible data foundation for advancing research in multimodal agent tuning.  

We compare MMAT-1M with several similar agent tuning and CoT datasets, including LLaVA-Plus-v1~\cite{liu2024llava}, Visual CoT~\cite{shao2024visual}, LLaVA-CoT~\cite{xu2024llavacot} and MM-Traj~\cite{gao2024multi}, as shown in Table~\ref{tab:dataset_comparison}. It is evident that the scale of our dataset substantially exceeds that of comparable datasets. Furthermore, our dataset is equipped with API and RAG tool invocation capabilities, supports CoT reasoning and reflection, and encompasses both one-turn and multi-turn reasoning paradigms.

\subsection{Data Engine}\label{sec:data_engine}
As shown in Figure~\ref{fig:data_engine_figure}, the data construction process is structured into four key stages: foundation, rationale generation, reflection, and integration of trajectories.

\noindent\textbf{Foundation.} As an illustrative example, we randomly select an image and its corresponding question-answer pair from the original dataset. To ensure consistency in response styles across different datasets, we optimize the phrasing of the questions. For samples with shorter answers, we append a response style constraint at the end of the question, while keeping the original answer unchanged. Additionally, we prepare external tools for invocation, including Image Caption, OVD, OCR, Face Detection, and RAG. The Image Caption operator generates textual descriptions of images, extracting key visual information and expressing their semantics. Based on the CCoT~\cite{mitra2024compositional}, we use GPT-4o to construct a scene graph and generate image descriptions accordingly, enhancing semantic understanding and compositional reasoning capabilities.
OVD leverages object information from the scene graph to identify and detect targets within an open vocabulary range, enabling the recognition of novel categories that extend beyond a predefined label set. This functionality is implemented using Grounding DINO~\cite{liu2024grounding}. OCR utilizes PaddleOCR~\cite{du2020pp} to recognize textual content within images. Face Detection, powered by deepface~\cite{serengil2024benchmark}, accurately locates facial regions in images.
Finally, for questions that require online search capabilities, we leverage GPT-4o to generate search queries, which are then used to invoke the Google API to retrieve the top-k most relevant information.

\noindent\textbf{Rationale.} We employ an iterative diagram to generate rationales, where the annotation process is powered by GPT-4o, ensuring the stability and efficiency of reasoning. During inference, the model adaptively invokes multimodal operators RAG to maintain the completeness and interpretability of the reasoning chain. The reasoning process initiates with problem analysis, where the model selects appropriate operators based on task requirements. If holistic semantic understanding is required, the Image Caption operator is invoked to extract a scene graph and generate an image description. For tasks demanding object-level information, the OVD operator is utilized to identify objects within an open-vocabulary range. Similarly, the OCR operator and Face Detection operator are employed for text recognition and facial analysis, respectively. When operator outputs are insufficient to support inference, the model formulates RAG queries to retrieve and integrate external knowledge.
Each reasoning step is meticulously recorded in a structured STRING format, capturing inference thoughts, operator invocations, retrieval requests, and subsequent actions. This adaptive multi-turn reasoning mechanism ensures the logical coherence of the reasoning chain, ultimately producing accurate, interpretable, and well-documented rationales.

\noindent\textbf{Reflection.} In our observations, the rationales generated through the process mentioned above exhibit two notable issues. The first is incompleteness in the reasoning process, particularly evident in the derivation of mathematical problems. This occurs when certain steps are omitted, making it challenging to arrive at the final answer. The second issue is reasoning cheating behavior, where the rationale's thought process does not logically lead to the final answer, but GPT-4o forcibly aligns the reasoning with the answer during label generation, creating an illusion of correctness.
To address these issues, we introduce reflective steps aimed at enhancing the model's error-correction capabilities during training and ensuring the reasoning process remains logically sound. Specifically, for the first issue, GPT-4o is tasked with identifying whether ``step skipping'' behavior exists in the reasoning process. If such behavior is detected, missing steps are supplemented to complete the derivation. For the second issue, we employ GPT-4o to re-evaluate whether the rationale's thought process aligns with the final answer. If a mismatch is identified, a reflective process is implemented to make the rationale aware of the cheating behavior and correct it accordingly.

\noindent\textbf{Integration.} The dataset generated through the approach above adopts a multi-turn Rationale and Reflection (RR) format, which may be impractical for real-world applications requiring time-sensitive responses. Inspired by the LUMOS~\cite{yin2024agent} model, we aim to create a dataset where the model can deliberate and produce the final answer in one turn. However, due to the constraints of the one-turn format, we cannot dynamically incorporate the results of external operators during the output phase. To address this, we integrate the results of all operators (excluding RAG) into the input stage, clearly demarcated by brackets. At the output stage, we consolidate multiple trajectories from the multi-turn dialogue into a One-turn Rationale and Reflection (ORR) format. Our findings indicate that ORR not only retains the ability to perform reasoning and integrate external tool results but also significantly improves inference speed, making it more suitable for time-critical applications.

To assess potential GPT-4o hallucinations, we evaluated all MMAT-1M samples on coherence, relevance, accuracy, completeness, and image-text alignment, with over 89\% demonstrating high-quality reasoning. Evaluation criteria are detailed in the supplementary material.

\subsection{Multimodal Agent Tuning}\label{sec:mmat}
Given a training sample: \(\{\{q_1, r_1\}, ...\{q_i, r_i\},...\{q_n, A\}\}\), where \(q_i\) is i-th question, \(r_i\) indicates the rationale, and A signifies the final answer. We select several open-source multimodal models and employ supervised fine-tuning (SFT) training schemes on these models.

\noindent\textbf{SFT.} We opt for low-rank adaptation (LoRA)~\cite{hu2022lora}, compared to full parameters fine-tuning, which not only retains the majority of the baseline model's knowledge but also save memory and computational space efficiently. The loss function of it is designed as follows:
\begin{equation}
L = L_{\text{original}} + \lambda \sum_{i} \| \Delta \theta_i \|_F^2,
\end{equation}
where \( L_{\text{original}} \) is the original loss function, \( \Delta \theta_i \) indicates the update of the \(i\)-th weight matrix, \( \lambda \) is the regularization parameter, and \( \| \cdot \|_F \) denotes the Frobenius norm.

\section{Experiments}
We conduct extensive experiments across multiple benchmarks to evaluate the effectiveness of our approach. Section~\ref{sec:implementation} details the implementation settings. In Section~\ref{sec:main_results}, we compare our method, which fine-tunes MLLMs with One-turn Rationale and Reflection (ORR) and Rationale and Reflection (RR) strategies on the MMAT-1M dataset, against baselines. The evaluation spans eight benchmarks, covering general and reasoning tasks, along with one benchmark for external knowledge retrieval. Section~\ref{sec:analysis} presents ablation studies and analyzes inference efficiency. Finally, Section~\ref{sec:qual_results} provides qualitative results for further insights into our method.
\begin{table*}[t!]
\footnotesize
\centering
\resizebox{\textwidth}{!}{
\begin{tabular}{llccccccccc}
\toprule
Model & Method & Average & MMStar & MMMU & MathVista & MathVision & AI2D & OCRBench & RealWorldQA & HallusionBench \\ 
\midrule
GPT-4o~\cite{hurst2024gpt} & \multicolumn{1}{c}{/} & 65.6 & 65.1 & 70.7 & 60.0 & 30.4 & 84.9 & 806 & 76.5 & 56.2 \\
\midrule
\multirow{3}{*}{Llama-3.2-11B-Vision-Instruct~\cite{meta2024llama}} 
& Baseline & 52.2 & 47.7 & 50.3 & 48.0 & 16.4 & 77.1 & 756 & 63.4 & 39.4 \\ 
& ORR & 54.6 & 50.7 & 47.8 & \textbf{50.1} & \textbf{17.7} & \textbf{78.9} & \textbf{806} & 66.7 & 44.4\\
& RR & \textbf{55.3} & \textbf{51.4} & \textbf{51.0} & 49.1 & 16.8 & 77.9 & 784 & \textbf{69.3} & \textbf{48.3} \\
\midrule
\multirow{3}{*}{MiniCPM-V-2.6~\cite{yao2024minicpm}}
& Baseline & 58.0 & 56.5 & 47.1 & 60.3 & 22.4 & 81.5 & 843 & 65.0 & 47.1 \\
& ORR & 58.8 & 56.9 & 47.9 & 60.6 & 23.4 & 81.7 & \textbf{848} & 66.6 & 48.8\\
& RR & \textbf{59.9} & \textbf{58.5} & \textbf{49.2} & \textbf{61.9} & \textbf{25.3} & \textbf{82.0} & 840 & \textbf{68.0} & \textbf{50.0} \\
\midrule
\multirow{3}{*}{InternVL2.5-2B~\cite{chen2024expanding}}
& Baseline & 52.7 & 53.6 & 43.2 & 50.1 & 16.1 & 75.1 & 804 & 60.5 & 42.6 \\
& ORR & 54.4 & \textbf{55.4} & \textbf{44.7} & 50.1 & 14.1 & \textbf{77.5} & \textbf{819} & \textbf{69.5} & 42.4 \\
& RR & \textbf{54.7} & 54.9 & 44.4 & \textbf{52.6} & \textbf{16.5} & 77.2 & 799 & 68.0 & \textbf{43.8} \\
\midrule
\multirow{3}{*}{InternVL2.5-4B~\cite{chen2024expanding}}
& Baseline & 58.4 & 58.6 & 51.8 & 60.8 & 21.7 & 81.2 & 823 & 64.6 & 46.5 \\
& ORR & 59.5 & 59.2 & 50.7 & 61.4 & 19.7 & 81.4 & \textbf{824} & 69.2 & \textbf{51.9} \\
& RR & \textbf{60.6} & \textbf{60.9} & \textbf{53.1} & \textbf{62.0} & \textbf{22.4} & \textbf{82.7} & 805 & \textbf{72.2} & 50.7 \\
\midrule
\multirow{3}{*}{InternVL2.5-8B~\cite{chen2024expanding}}
& Baseline & 60.7 & 62.4 & 53.1 & 64.5 & 20.1 & 84.1 & 819 & 69.4 & 49.8 \\
& ORR & 62.4 & 64.8 & 55.4 & 63.8 & 20.8 & 83.5 & \textbf{849} & 73.0 & 53.3 \\
& RR & \textbf{63.4} & \textbf{65.3} & \textbf{57.3} & \textbf{64.8} & \textbf{21.7} & \textbf{84.2} & 839 & \textbf{74.4} & \textbf{55.8} \\
\bottomrule
\end{tabular}
}
\caption{Performance comparison of MLLMs with Baseline, ORR (One-turn Rationale and Reflection), and RR (Rationale and Reflection) across eight benchmarks. Models trained on MMAT-1M with ORR and RR achieve overall gains, enhancing multimodal capabilities.}
\label{tab:bench}
\end{table*}

\begin{table}[h!]
\setlength{\belowcaptionskip}{-0.6cm}
\footnotesize
\centering
\begin{tabular}{lcc}
\toprule
Model & Query & Golden Query \\
\midrule
GPT-4o~\cite{hurst2024gpt} & 52.0 & 61.5 \\
OmniSearch (GPT-4V)~\cite{li2024benchmarking} & 50.0 & / \\
\midrule
Llama-3.2-11B-Vision-Instruct~\cite{meta2024llama} & 29.4 & 34.6 \\
Llama-3.2-11B-Vision-Instruct-RR & \textbf{38.0} & \textbf{45.1} \\
\midrule
MiniCPM-V-2.6~\cite{yao2024minicpm} & 32.7 & 39.2 \\
MiniCPM-V-2.6-RR & \textbf{35.9} & \textbf{44.4} \\
\midrule
InternVL2.5-2B~\cite{chen2024expanding} & 19.3 & 26.0 \\
InternVL2.5-2B-RR & \textbf{30.9} & \textbf{38.8} \\
\midrule
InternVL2.5-4B~\cite{chen2024expanding} & 23.3 & 31.1 \\
InternVL2.5-4B-RR & \textbf{35.4} & \textbf{42.1} \\
\midrule
InternVL2.5-8B~\cite{chen2024expanding} & 27.0 & 35.2 \\
InternVL2.5-8B-RR & \textbf{36.8} & \textbf{44.0} \\
\bottomrule
\end{tabular}
\caption{Results on the RAG Benchmark Dyn-VQA. RR strategy significantly boosts performance across model scales, enhancing multi-hop reasoning and retrieval.}
\label{tab:dyn_vqa}
\end{table}

\subsection{Implementation Details}\label{sec:implementation}
In this section, we integrate MMAT-1M with various MLLMs to showcase the broad applicability of our approach. We investigate two reasoning strategies, ORR and RR, which guide multimodal models toward structured and interpretable reasoning. ORR consolidates all reasoning steps into a single query, enabling efficient inference while maintaining strong accuracy. In contrast, RR follows a multi-step reasoning process, dynamically selecting operators and retrieving external knowledge when needed. For reasoning scenarios that require external knowledge injection, we employ Google Search to retrieve relevant information. Each query returns up to three results (top-k=3), providing the model with necessary contextual knowledge while maintaining efficiency. 

We apply these strategies to open-source multimodal models, including Llama-3.2-11B-Vision-Instruct~\cite{meta2024llama}, MiniCPM-V-2.6~\cite{yao2024minicpm}, and the InternVL2.5 series~\cite{chen2024expanding}, which includes InternVL2.5-2B, InternVL2.5-4B, and InternVL2.5-8B. Each model is separately fine-tuned with ORR and RR on the MMAT-1M dataset, which consists of 1,090,263 question-answer pairs, for one epoch with a learning rate of 4e-5. Detailed training parameters are provided in the supplementary material.

\subsection{Main Results on Benchmark}\label{sec:main_results}
\begin{table*}[h!]
\centering
\resizebox{\textwidth}{!}{
\begin{tabular}{lcc|cccccccccc}
\toprule
Model & API & RAG & Average & MMStar & MMMU & MathVista & MathVision & AI2D & OCRBench & RealWorldQA & HallusionBench & Dyn-VQA \\ 
\midrule
Baseline & \scalebox{0.75}{\usym{2613}} & \scalebox{0.75}{\usym{2613}} & 57.9 & 62.4 & 53.1 & 65.1 & 20.1 & 84.1 & 819 & 69.4 & 49.8 & 35.2 \\
\midrule
Baseline-RR & \checkmark & \scalebox{0.75}{\usym{2613}} & 59.8 & 65.0 & 56.2 & 64.2 & 20.4 & 84.1 & 839 & 74.3 & 55.0 & 35.4 \\  
Baseline-RR & \scalebox{0.75}{\usym{2613}} & \checkmark &  57.3 & 60.1 & 52.6 & 61.1 & 21.0 & 81.8 & 797 & 67.8 & 48.0 & 43.4 \\  
Baseline-RR (w/o SFT) & \checkmark & \checkmark & 55.0 & 60.6 & 49.8 & 60.9 & 15.1 & 82.8 & 825 & 68.9 & 43.2 & 31.5 \\ 
Baseline-R & \checkmark & \checkmark & 60.2 & 65.0 & 54.5 & 63.9 & 20.5 & 84.6 & 826 & 72.7 & 54.8 & 42.9 \\ 
Baseline-ORR & \checkmark & \scalebox{0.75}{\usym{2613}} & 59.6 & 64.8 & 55.4 & 63.8 & 20.8 & 83.5 & \textbf{849} & 73.0 & 53.3 & 36.6 \\ 
Baseline-RR & \checkmark & \checkmark & \textbf{61.3} & \textbf{65.3} & \textbf{57.3} & \textbf{64.8} & \textbf{21.7} & \textbf{84.2} & 839 & \textbf{74.4} & \textbf{55.8} & 44.0 \\ 
\bottomrule
\end{tabular}
}
\caption{Ablation study evaluating the impact of SFT, API integration, structured reflection, and RAG-based retrieval on multimodal reasoning performance. Results highlight the complementary benefits of fine-tuning, explicit rationale generation, and external knowledge integration in enhancing multimodal reasoning performance.}
\label{tab:ablation}
\end{table*}
\noindent\textbf{Setup.} We conduct a comprehensive evaluation of our method using eight widely adopted and challenging benchmarks: MMStar~\cite{chenwe}, MMMU~\cite{yue2024mmmu}, MathVista~\cite{lumathvista}, MathVision~\cite{wang2025measuring}, AI2D~\cite{kembhavi2016diagram}, OCRBench~\cite{liu2023hidden}, RealWorldQA~\cite{grok15}, and HallusionBench~\cite{guan2024hallusionbench}. Specifically, MMStar and MMMU primarily assess multimodal reasoning and question-answering capabilities, while MathVista and MathVision focus on mathematical and visual reasoning skills. AI2D examines the comprehension of scientific diagrams, and OCRBench evaluates textual information extraction from documents. RealWorldQA targets spatial reasoning in real-world scenarios, whereas HallusionBench gauges susceptibility to language hallucinations and visual illusions. For MathVista and MathVision, we adopt the testmini set. To ensure fairness and reproducibility, all evaluations are conducted using VLMEvalKit~\cite{duan2024vlmevalkit}, an open-source toolkit specifically designed for large vision-language models.
Beyond these benchmarks, we further evaluate the RAG capabilities of the models with the Dyn-VQA dataset proposed in OmniSearch~\cite{li2024benchmarking}. Dyn-VQA encompasses dynamic, multimodal, multi-hop reasoning tasks, offering a comprehensive assessment of how effectively models plan retrieval strategies and integrate relevant information.

\noindent\textbf{Main Results.} Table~\ref{tab:bench} presents experimental results on multiple benchmarks that evaluate the performance of various multimodal large models trained on MMAT-1M using ORR and RR. The findings demonstrate that both methods effectively enhance model performance across different parameter scales.

Training with our ORR on MMAT-1M improves the average score of InternVL2.5-8B from 60.7 to 62.4 compared to the baseline, while our RR strategy further boosts it to 63.4. Notably, RR consistently outperforms the baseline and achieves competitive results against GPT-4o. Specifically, InternVL2.5-8B with RR surpasses GPT-4o on MMStar (65.3 vs. 65.1) and MathVista (64.8 vs. 60.0), demonstrating superior multimodal reasoning and mathematical-visual understanding. It also outperforms GPT-4o on OCRBench (839 vs. 806), reflecting stronger textual information extraction. Additionally, it performs on par with GPT-4o on AI2D (84.2 vs. 84.9) and HallusionBench (55.8 vs. 56.2), indicating robust comprehension of scientific diagrams and resilience to multimodal hallucinations. 

Compared with baseline models such as InternVL2.5-8B, MiniCPM-V-2.6, and Llama-3.2-11B-Vision-Instruct, our ORR and RR particularly RR, have demonstrated generally similar optimization effects across various test sets. Our RR on MiniCPM-V2.6 achieves a gain in average from 58.0 to 59.9, a 3.3\% relative increase, while on Llama-3.2-11B-Vision-Instruct achieves a gain from 52.2 to 55.3, a relative improvement of 5.9\%. This indicates that our methods have broad applicability across different model series.
Similarly, our ORR and RR consistently deliver strong performance across the InternVL2.5 series, including the 2B, 4B, and 8B parameter variants, demonstrating robust scalability and wide-ranging applicability of our methodology. 

In OCRBench, InternVL2.5-2B's ORR strategy outperforms the baseline (804 to 819), while RR drops to 799, a trend also seen in InternVL2.5-4B and 8B. The reason for this phenomenon is that, although RR exhibits specific error-correction capabilities, the OCR misrecognition negatively impacts the final results. In contrast, ORR utilizes image captioning to mitigate OCR errors, demonstrating superior performance in OCRBench. 

The comprehensive results confirm that training on MMAT-1M with our ORR and RR leads to significant improvements, particularly with RR, in tasks requiring comprehensive reasoning, mathematical computation, and cross-modal information fusion. This establishes MMAT-1M as a valuable benchmark for advancing the reasoning capabilities of vision-language models.

\noindent\textbf{Results on RAG Benchmark. } The evaluation results of Dyn-VQA~\cite{li2024benchmarking} are shown in Table~\ref{tab:dyn_vqa}, based on the latest version. Query refers to the input content used by the model for information retrieval, while Golden Query denotes an optimized prompt focused on the final retrieval step to maximize answer accuracy. To align with Dyn-VQA, we adopt the same evaluation metric, F1-Recall, which measures the overlap between the model-generated response and the ground truth. Results demonstrate that our ORR and RR consistently enhance multi-hop reasoning and retrieval performance. Specifically, the RR improves Llama-3.2-11B-Vision-Instruct by 29.3\% relative to its original performance 
 (from 29.4 to 38.0) in Query and by 30.3\% relative to its original performance (from 34.6 to 45.1) in Golden Query, while MiniCPM-V-2.6 shows improvements of 9.8\% and 13.3\%, respectively. The InternVL2.5 series models similarly benefit, with relative gains ranging from 31.9\% to 60.1\%, underscoring the effectiveness of our methods across complex, knowledge-intensive tasks.

\begin{figure}[h!]
    \centering
    \includegraphics[width=0.9\linewidth]{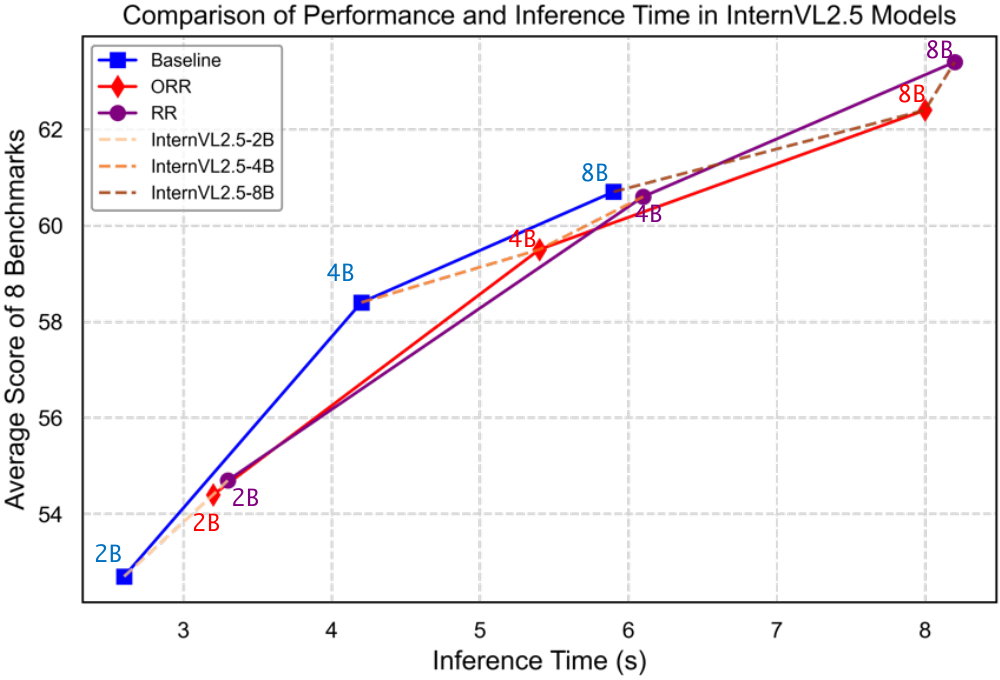}
    \caption{Comparison of inference efficiency and performance gains of ORR and RR across different InternVL2.5 model scales.}
    \label{fig:inference_time}
\end{figure}

\begin{figure*}[htbp]
    \centering
    \includegraphics[width=0.82\textwidth]{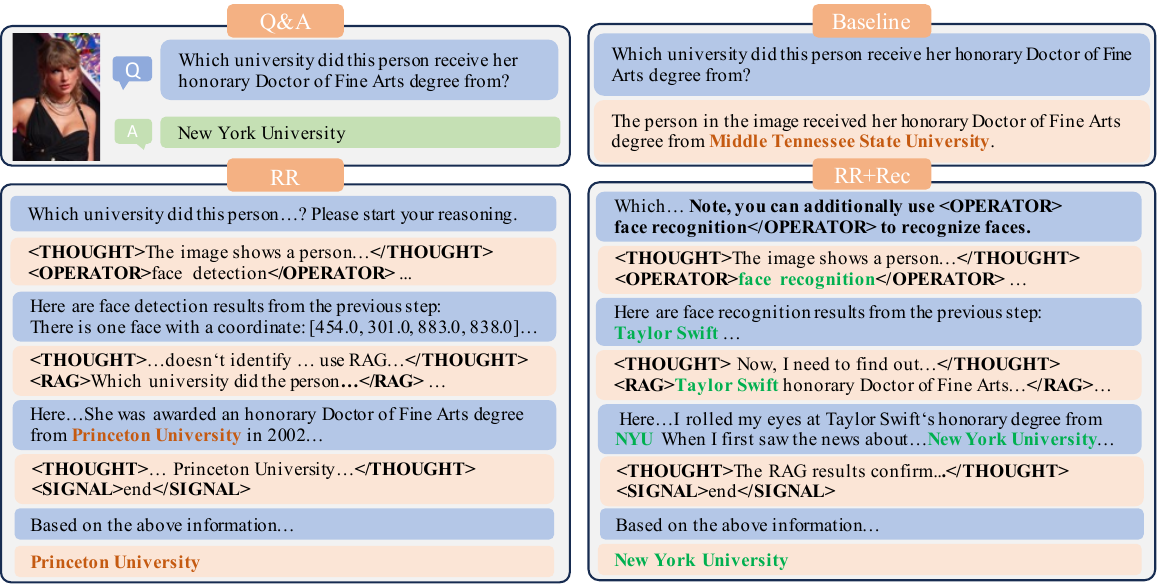}
    \caption{The zero-shot capability of invoking a celebrity recognition operator of InternVL2.5-8B-RR.}
    \label{fig:face_recog}
\end{figure*}

\subsection{Further Analysis}\label{sec:analysis}
\noindent\textbf{Ablation Study. }Table~\ref{tab:ablation} presents an ablation study on the effects of SFT, API integration, structured reflection, and RAG on multimodal reasoning performance.
The baseline model, without external resources, achieves an average score of 57.9. In the RR setting, Baseline-RR achieves the highest score of 61.3 with both API and RAG. Removing API reduces performance to 57.3, while removing RAG lowers it to 59.8. Without SFT, performance declines further to 55.0. Additionally, Baseline-R, which retains rationale but omits reflection, scores 60.2, suggesting that reflection enhances reasoning ability. In the ORR setting, performance declines to 59.6, primarily because the ORR format does not incorporate RAG information, resulting in a performance drop on the Dyn-VQA benchmark. On other benchmarks, however, its performance remains comparable to that of the RR format. These results confirm that SFT is crucial for instruction adherence, while structured reflection and external knowledge integration further improve multimodal reasoning.

\noindent\textbf{Performance Efficiency Tradeoff between ORR and RR.} Figure~\ref{fig:inference_time} compares the inference efficiency and performance gains of the ORR and RR methods across different InternVL2.5 model scales. Although both ORR and RR consistently enhance multimodal reasoning performance, their inference times notably increase relative to the baseline. ORR introduces a moderate inference overhead due to its one-turn structured reasoning approach, while RR, involving multi-turn adaptive reasoning steps, incurs a slightly higher computational cost. However, RR achieves greater performance improvements compared to ORR, demonstrating a beneficial tradeoff between computational efficiency and reasoning accuracy.

\subsection{Qualitative Results}\label{sec:qual_results}
While the experiments, as mentioned above, have demonstrated the benefits of invoking external tools for the model, the capabilities of a fixed set of tools are inherently limited. For instance, MMAT-1M's lack of a celebrity recognition operator hinders the fine-tuned model from achieving correct results in cases requiring celebrity identification. To address this, we conduct an experiment to verify whether the fine-tuned model can invoke operators it has not been explicitly trained on. As shown in Figure~\ref{fig:face_recog} , we test a visual question with the InternVL2.5-8B model. Initially, the baseline model provides an incorrect answer. As anticipated, the model fine-tuned on MMAT-1M, failing to recognize the person, also returns a wrong answer due to unsuccessful web search results. To address this limitation, we instruct the fine-tuned model to invoke a celebrity recognition operator, which successfully identifies the correct answer. This experiment demonstrates that the model fine-tuned on our dataset exhibits a certain level of zero-shot capability for invoking unseen tools. However, its performance remains inferior to that achieved through explicit fine-tuning.

\section{Conclusion}
The introduction of MMAT-1M represents a significant advancement in multimodal agent tuning, offering a diverse and flexible dataset for enhancing CoT reasoning and tool usage in MLLMs. By addressing key limitations of existing multimodal agent tuning datasets, such as homogeneity, lack of reflection, and inflexible tool usage, it provides a comprehensive solution that aligns with the demands of real-world applications. While the dataset demonstrates robust performance on current multimodal benchmarks, further research is essential to evaluate its adaptability to a broader array of MLLMs and more intricate real-world scenarios.

{
    \small
    \bibliographystyle{ieeenat_fullname}
    \bibliography{main}
}

\clearpage
\appendix









\maketitlesupplementary

\renewcommand\thesection{\Alph{section}}

\section{Dataset Statistics}
Table~\ref{tab:data_category} presents detailed statistics of the datasets utilized to construct MMAT-1M. These datasets are curated from various prominent sources, including Visual CoT~\cite{shao2024visual}, LLaVA-CoT~\cite{xu2024llavacot}, The Cauldron~\cite{laurenccon2024matters}, TabMWP~\cite{lu2022dynamic}, and InfoSeek~\cite{chen2023can}, collectively contributing to an extensive multimodal reasoning dataset. Specifically, the table enumerates the composition, the number of data entries, and the corresponding QA pairs of each sub-dataset. In total, the MMAT-1M dataset comprises 1,090,263 QA pairs, indicating substantial coverage and diversity in multimodal tasks.

\begin{table}[htbp]
\footnotesize
\centering
\resizebox{\linewidth}{!}{
\begin{tabular}{lcc}
    \toprule
    Category & Number of Data Entries & Number of QA Pairs \\
    \midrule
    \multicolumn{3}{l}{Visual CoT~\cite{shao2024visual}} \\
    \quad Birds-200-2011~\cite{wah2011caltech} & 10.1k & 10.1k \\
    \quad DocVQA~\cite{mathew2021docvqa} & 33.5k & 33.5k \\
    \quad DUDE~\cite{van2023document} & 11.7k & 11.7k \\
    \quad Flickr30K~\cite{plummer2015flickr30k} & 135.7k & 135.7k \\
    \quad GQA~\cite{hudson2019gqa} & 98.1k & 98.1k \\
    \quad InfographicsVQA~\cite{mathew2022infographicvqa} & 15.1k & 15.1k \\
    \quad Open images~\cite{kuznetsova2020open} & 43.1k & 43.1k \\
    \quad SROIE~\cite{huang2019icdar2019} & 2.5k & 2.5k \\
    \quad TextCap~\cite{sidorov2020textcaps} & 32.2k & 32.2k \\
    \quad TextVQA~\cite{singh2019towards} & 18.5k & 18.5k \\
    \quad Visual7W~\cite{zhu2016visual7w} & 30.5k & 30.5k \\
    \quad VSR~\cite{Liu2022VisualSR} & 3.4k & 3.4k \\
    \midrule
    \multicolumn{3}{l}{LLaVA-CoT~\cite{xu2024llavacot}} \\
    \quad ShareGPT4V~\cite{sharegpt4v} & 31.3K & 67.9k \\
    \quad ChartQA~\cite{chartqa} & 17.0k & 25.6k \\
    \quad A-OKVQA~\cite{chartqa} & 16.1K & 99.9K \\
    \quad AI2D~\cite{kembhavi2016diagram} & 11.4k & 11.4k \\
    \quad GeoQA+~\cite{geoqa+} & 11.4k & 11.4k \\
    \quad ScienceQA~\cite{scienceqa} & 5.6k & 5.6k \\
    \quad DocVQA~\cite{mathew2021docvqa} & 4.0k & 31.3k \\ 
    \quad PISC~\cite{pisc} & 0.9k & 0.9k \\
    \quad CLEVR~\cite{clevr} & 0.5k & 0.5k \\ 
    \quad CLEVR-Math~\cite{clevr-math} & 0.5k & 0.5k \\
    \midrule
    \multicolumn{3}{l}{The Cauldron~\cite{laurenccon2024matters}} \\
    \quad HatefulMemes~\cite{hatefulmeme} & 8.5k & 8.5k \\
    \quad Screen2Words~\cite{screen2words} & 15.7k & 15.7k  \\
    \quad ST-VQA~\cite{STVQA} & 17.2k & 23.1k \\
    \quad VisText~\cite{VisText} & 10.0k & 10.0k \\
    \quad WikiSQL~\cite{WikiSQL} & 75.0k & 86.2k \\
    \quad WTQ~\cite{WTQ} & 38.2k & 44.1k \\
    \quad IconQA~\cite{IconQA} & 27.3k & 29.8k \\
    \quad RAVEN~\cite{RAVEN} & 20.9k & 20.9k \\
    \quad Inter-GPS~\cite{Inter-GPS} & 1.3k & 1.8k \\
    \quad TQA~\cite{TQA} & 1.5k & 6.5k \\
    \midrule
    \multicolumn{3}{l}{TabMWP~\cite{lu2022dynamic}} \\
    \quad TabMWP~\cite{lu2022dynamic} & 23.1k & 23.1k \\
    \midrule
    \multicolumn{3}{l}{InfoSeek~\cite{chen2023can}} \\
    \quad InfoSeek~\cite{chen2023can} & 131.4k & 131.4k \\
    \bottomrule
\end{tabular}
}
\caption{Detailed statistics of datasets included in MMAT-1M.}
\label{tab:data_category}
\end{table}

\section{Data Engine}
This section introduces the API operators, including scene graph-based image caption generation. It then details the prompts and their designs underlying the data engine, clearly distinguishing between rationale generation and reflection components.

\begin{figure}[htbp]
    \centering
    \includegraphics[width=\linewidth]{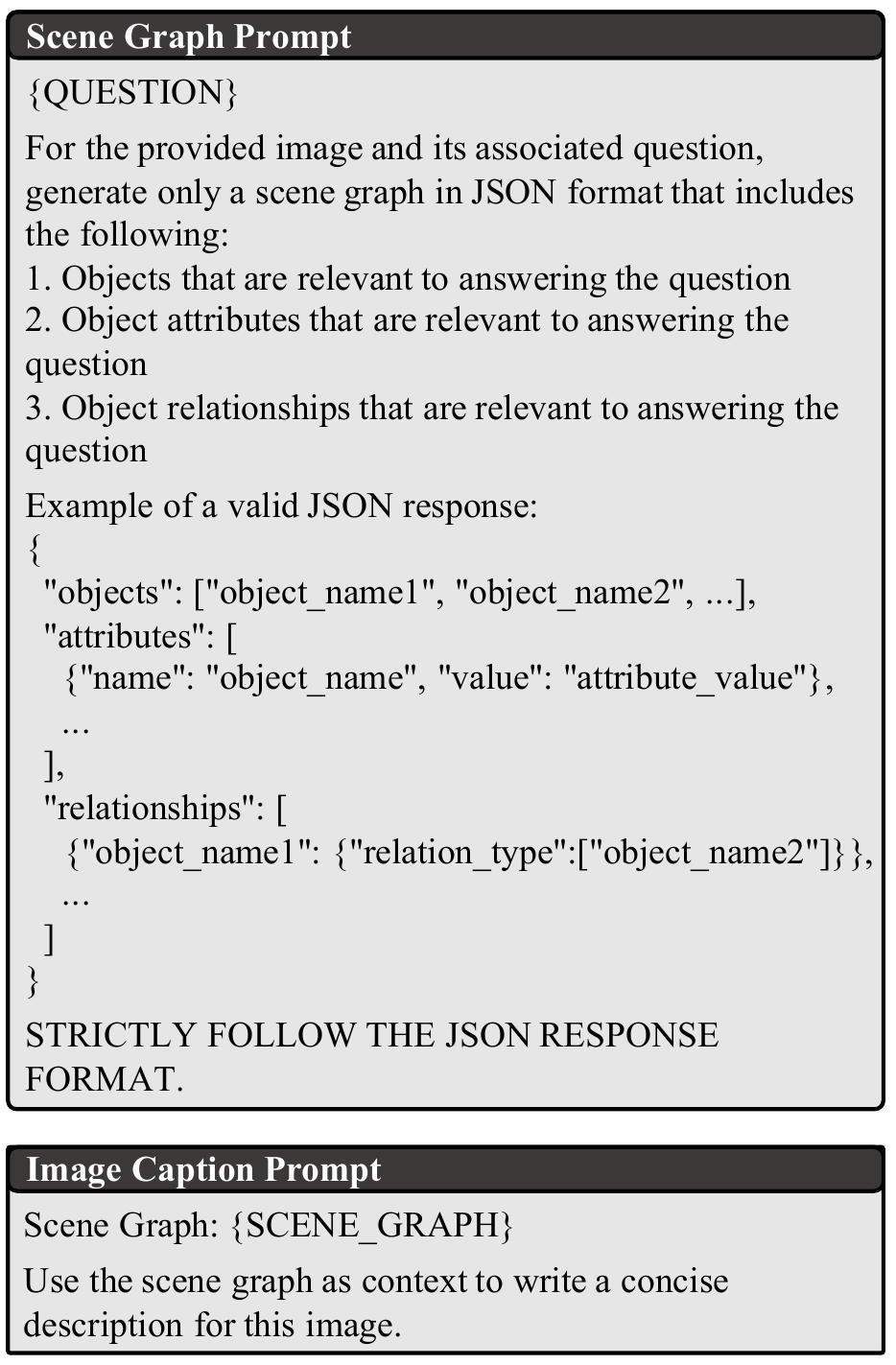}
    \caption{Scene graph and caption generation prompts.}
    \label{fig:scene_desc}
\end{figure}

Inspired by the CCoT~\cite{mitra2024compositional}, GPT-4o is leveraged to construct a scene graph and derive an image caption, enhancing semantic understanding and compositional reasoning. The scene graph prompt and image caption prompt are depicted in ~\ref{fig:scene_desc}.

\begin{figure}[htbp]
    \centering
    \includegraphics[width=\linewidth]{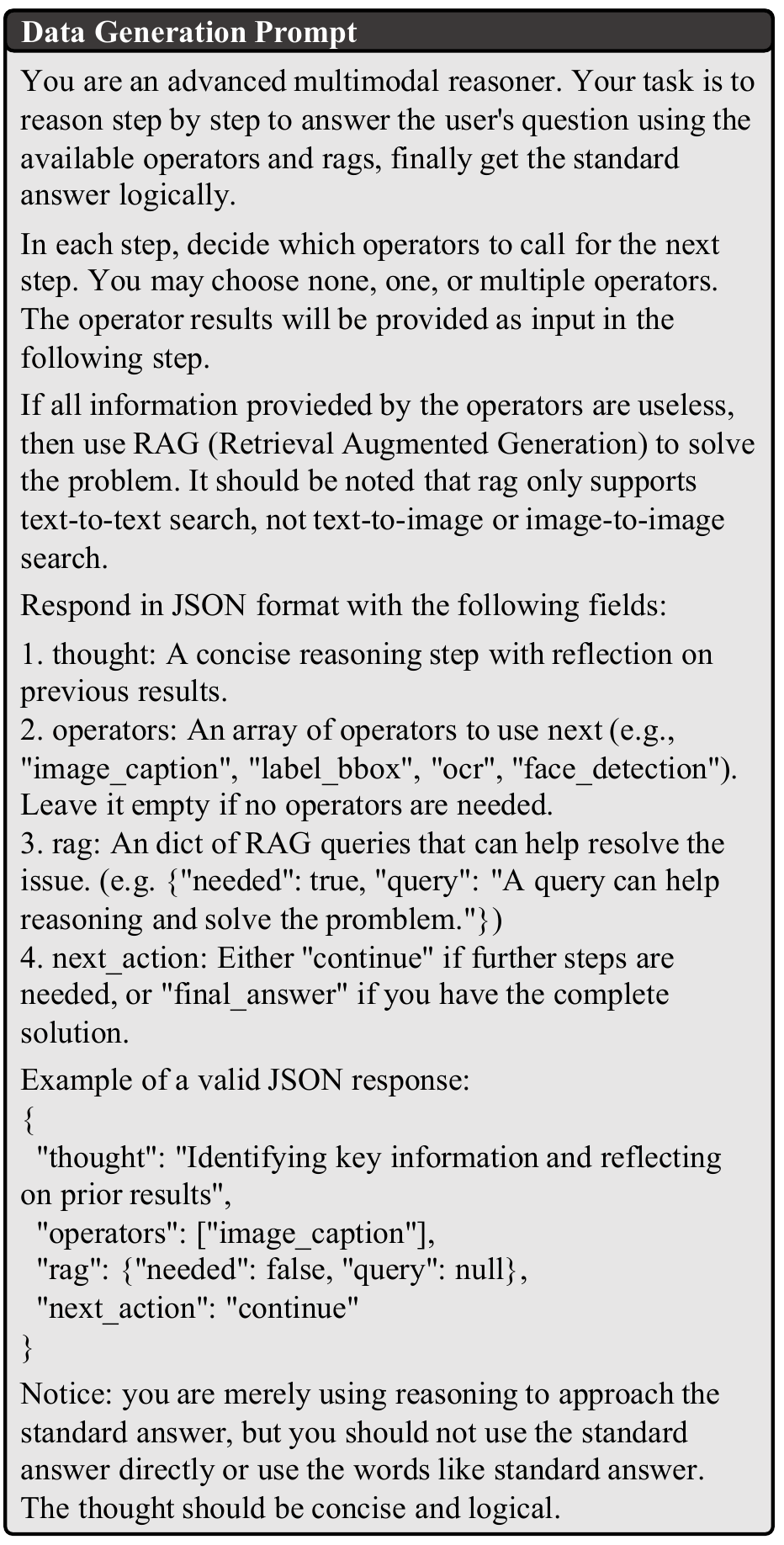}
    \caption{Rationale generation prompt.}
    \label{fig:data_generation}
\end{figure}

For rationale generation, we guide GPT-4o through a structured and adaptive multi-stage reasoning process. During inference, the model dynamically invokes multimodal operators, including Image Caption, Open-Vocabulary Objection Detection (OVD), Optical Character Recognition (OCR), and Face Detection, while also leveraging retrieval-augmented generation when needed. Each reasoning step is explicitly documented in structured JSON format to maintain transparency and logical coherence. Figure~\ref{fig:data_generation} illustrates the comprehensive design of the rationale generation prompt.

\begin{figure}[htbp]
    \centering
    \includegraphics[width=\linewidth]{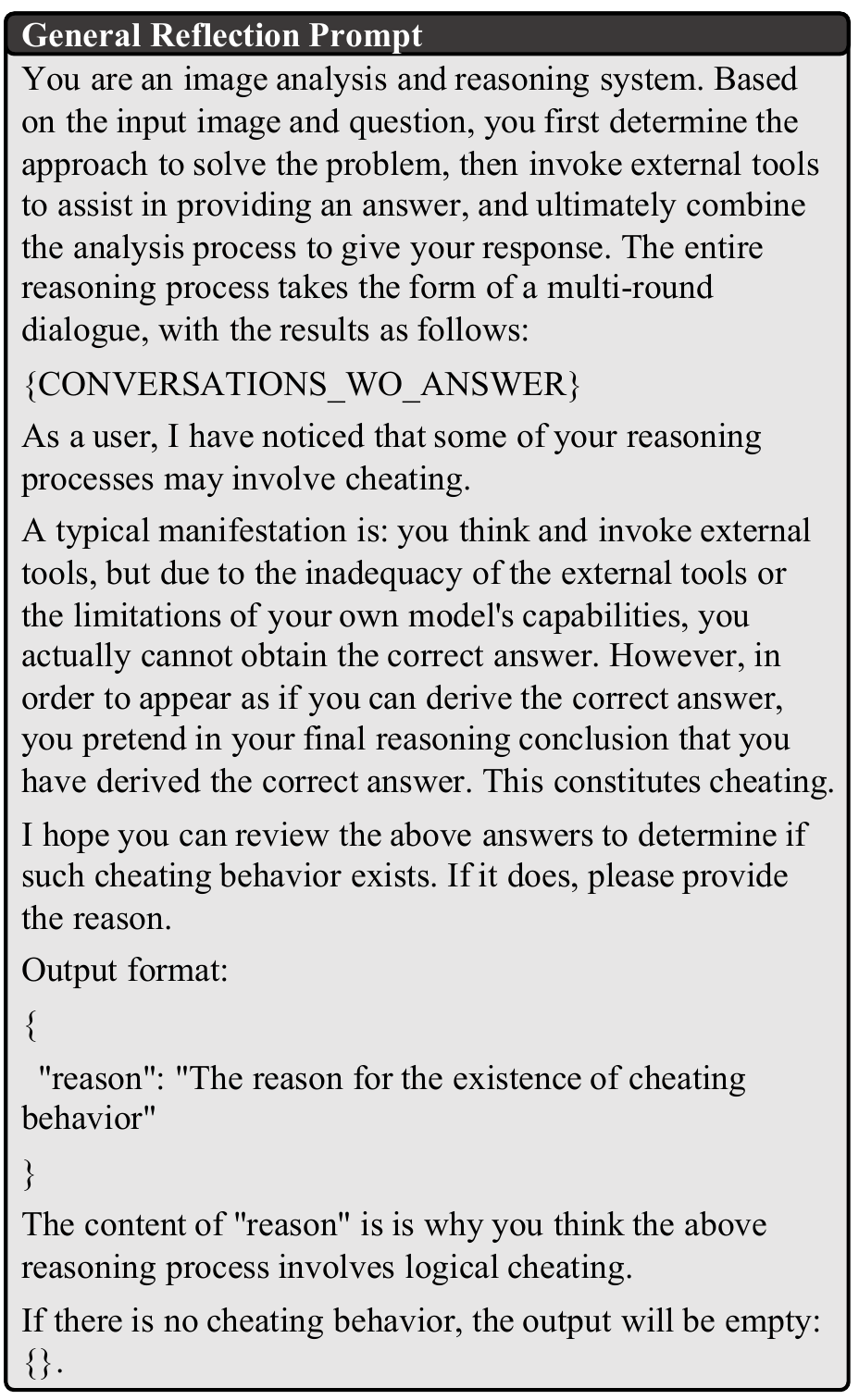}
    \caption{General reflection prompt.}
    \label{fig:general_reflection}
\end{figure}

For reflection, we designed two targeted prompts to enhance reasoning robustness. The general reflection prompt is intended to detect and correct reasoning cheating behaviors. Specifically, it prompts GPT-4o to critically examine cases where its reasoning process artificially aligns with given answers rather than deriving them through genuine inference. The prompt explicitly requests that the model identify and articulate any logical inconsistencies in its reasoning. The detailed structure of the general reflection prompt is provided in Figure~\ref{fig:general_reflection}.

\begin{figure}[htbp]
    \centering
    \includegraphics[width=\linewidth]{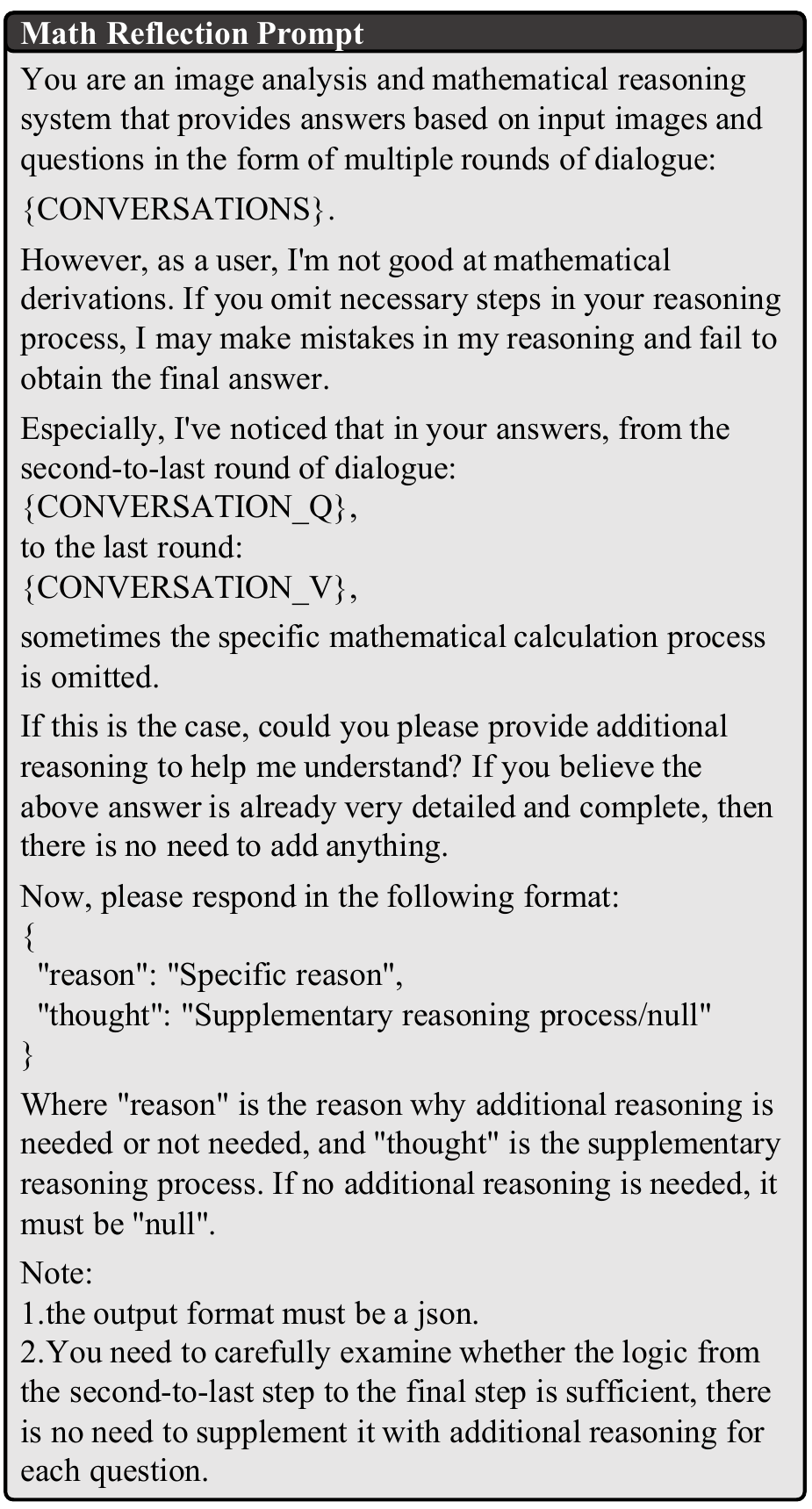}
    \caption{Math reflection prompt.}
    \label{fig:math_reflection}
\end{figure}

The math reflection prompt specifically targets completeness issues in mathematical reasoning. It instructs GPT-4o to carefully inspect its mathematical derivations, identifying instances where crucial calculation steps might be omitted. By prompting the model to supplement missing derivations explicitly, this ensures the integrity and clarity of mathematical reasoning. Figure~\ref{fig:math_reflection} depicts the math reflection prompt’s structure in detail. 

Our MMAT-1M dataset supports both one-turn and multi-turn reasoning frameworks, each with a dedicated system prompt.

\begin{figure}[h!]
    \centering
    \includegraphics[width=\linewidth]{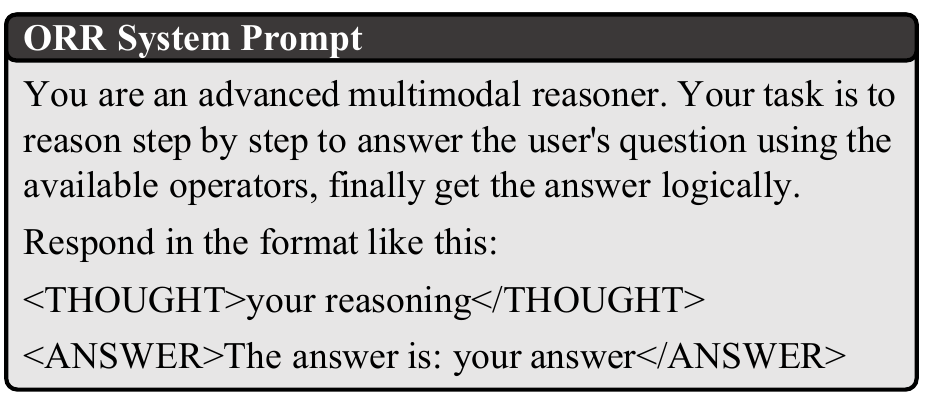}
    \caption{One-turn rationale and reflection (ORR) prompt.}
    \label{fig:orr_sys}
\end{figure}

The one-turn rationale and reflection (ORR) prompt enables the model to complete the entire reasoning process within a single inference step. This design ensures efficiency while maintaining strong reasoning capabilities. The structured output format ensures clarity and consistency. The specific prompt format for ORR is shown in Figure~\ref{fig:orr_sys}.

\begin{figure}[htbp]
    \centering
    \includegraphics[width=\linewidth]{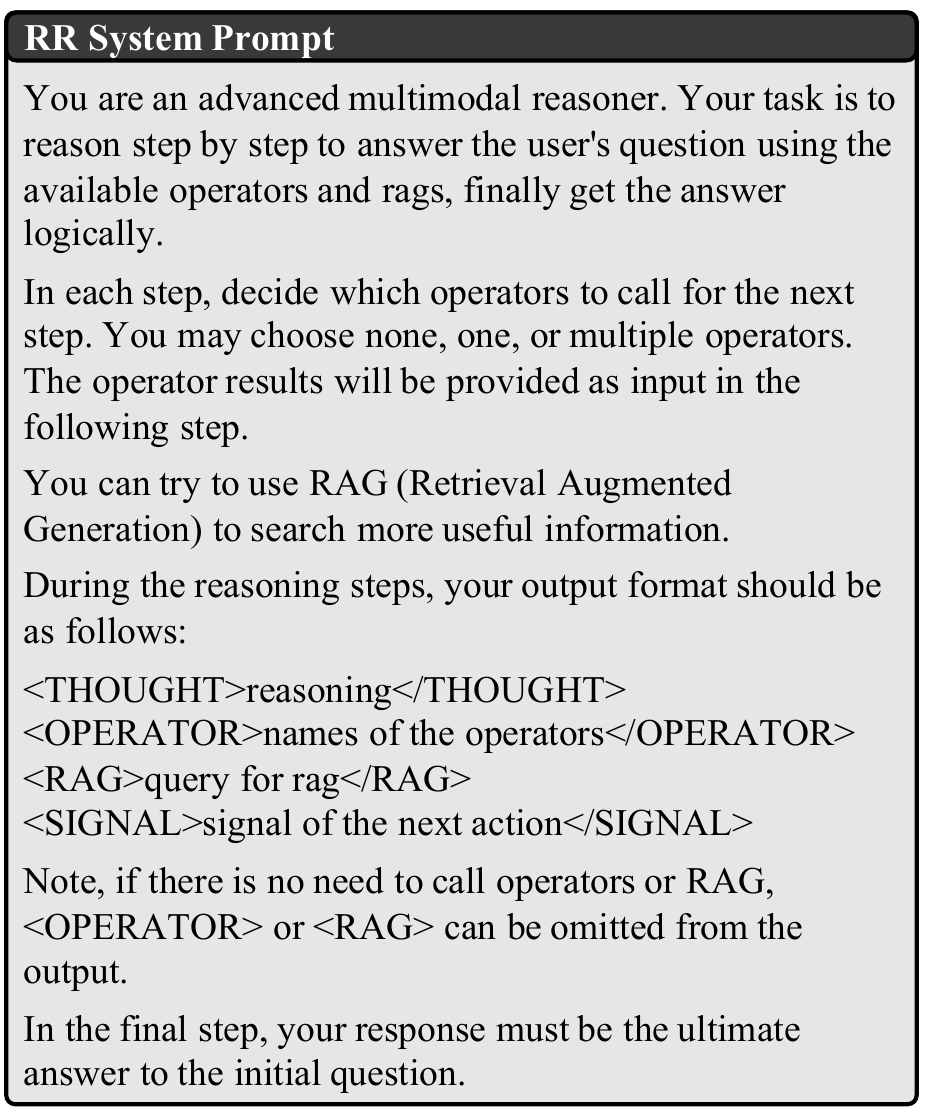}
    \caption{Rationale and reflection (RR) prompt.}
    \label{fig:rr_sys}
\end{figure}

The rationale and reflection (RR) prompt guides the model through an iterative reasoning process, dynamically selecting multimodal operators and retrieving external knowledge when necessary. This approach enhances interpretability and reasoning depth. Figure~\ref{fig:rr_sys} outlines the specific format designed for RR.

\begin{figure}[htbp]
    \centering
    \includegraphics[width=\linewidth]{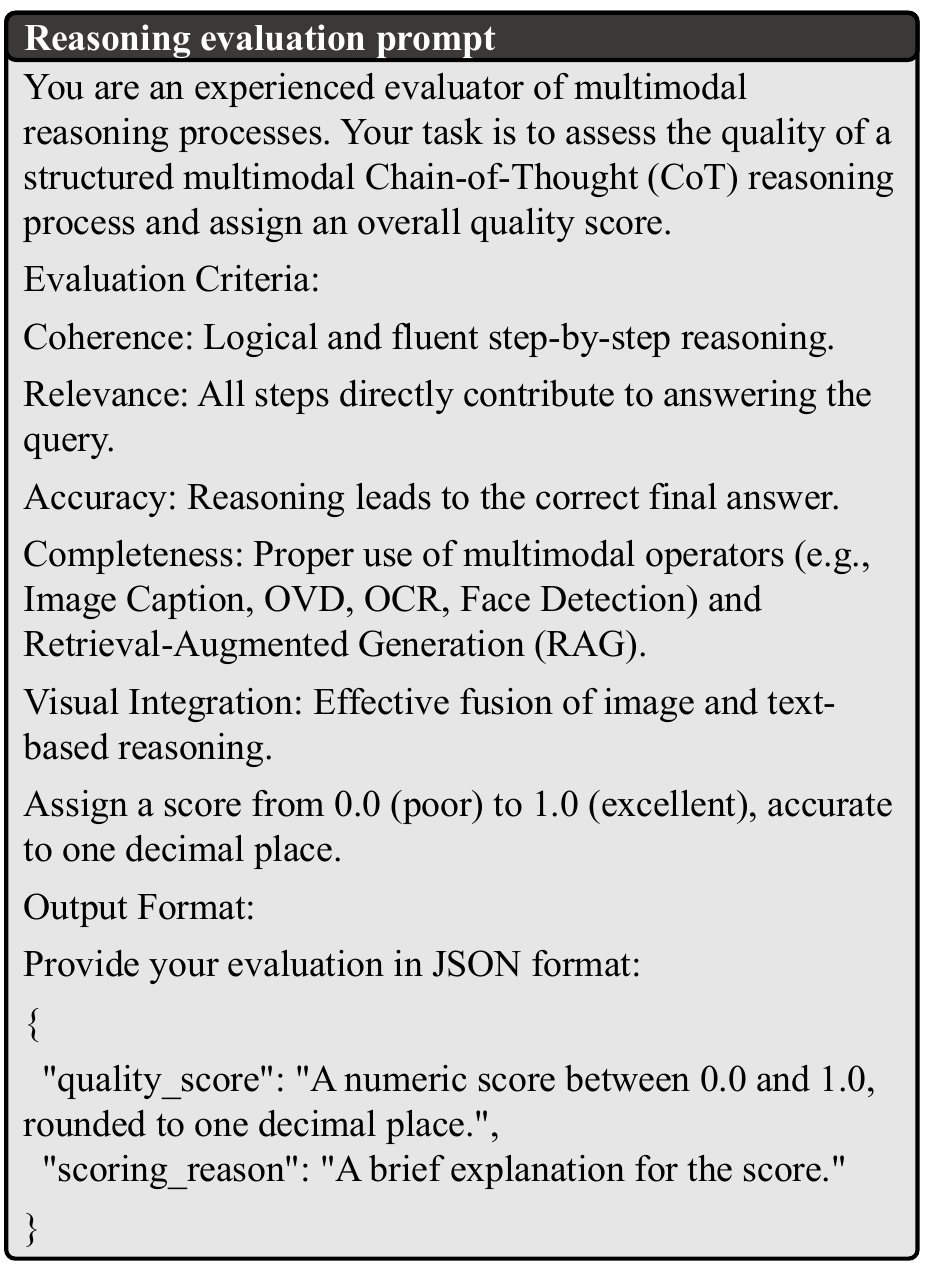}
    \caption{Reasoning evaluation prompt.}
    \label{fig:reasoning_eval}
\end{figure}

To assess potential GPT-4o hallucinations introduced during the rationale and reflection generation stages, we performed a large-scale quality evaluation of the entire MMAT-1M dataset. Specifically, we employed the Doubao-1.5-Vision-Pro-32K model to evaluate all samples against five criteria: coherence, relevance, accuracy, completeness, and image-text integration. Results indicate that over 89\% of the samples exhibit high-quality reasoning. Figure~\ref{fig:reasoning_eval} presents the prompt used in this evaluation.

\section{Training Hyperparameters}

In this section, we present the main training parameters for multiple models. For all models, including Llama-3.2-11B-Vision-Instruct~\cite{meta2024llama}, MiniCPM-V-2.6~\cite{yao2024minicpm}, and the InternVL2.5 series~\cite{chen2024expanding}, we adopt the same training configuration and use the open-source framework ms-swift~\cite{zhao2024swiftascalablelightweightinfrastructure} for training. The specific parameters are shown in Table~\ref{tab:hyperparameters}.
\begin{table}[htbp]
\small
\centering
\begin{tabular}{lc}
    \toprule
    Parameter & Value \\
    \midrule
    train\_type & LoRA \\
    num\_train\_epochs & 1 \\
    train\_batch\_size & 1 \\
    gradient\_accumulation\_steps & 1 \\
    learning\_rate & $4 \times 10^{-5}$ \\
    weight\_decay & 0.1 \\
    max\_length & 16384 \\
    torch\_dtype & BF16 \\
    seed & 42 \\
    deepspeed & ZeRO-2 \\
    \bottomrule
\end{tabular}
\caption{Configuration of hyperparameters used in training.}
\label{tab:hyperparameters}
\end{table}

\section{More Qualitative Examples}
In this section, we present additional qualitative examples highlighting the practical benefits of iterative reasoning, self-reflection, and one-turn reasoning.

\begin{figure*}[p]
    \centering
    \includegraphics[width=0.8\textwidth]{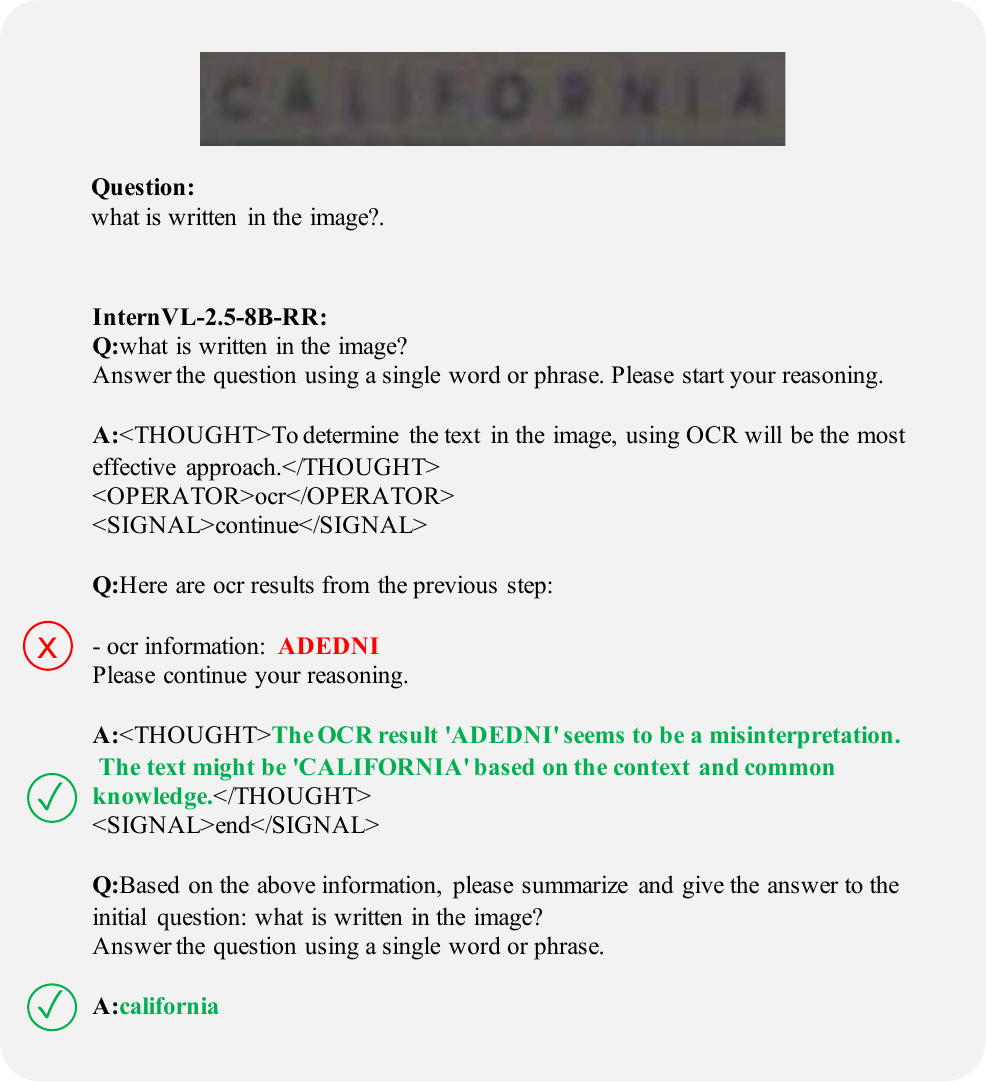}
    \caption{Example of iterative rationale result. The OCR recognition is erroneous, but the rationale process corrects the mistake.}
    \label{fig:RR_ocr_demo}
\end{figure*}

Figure~\ref{fig:RR_ocr_demo} illustrates how the model refines its reasoning to correct an OCR error. The initial OCR result misidentifies the text as ``ADEDNI'', but by leveraging contextual understanding and common knowledge, the model correctly recognizes it as ``CALIFORNIA''. This highlights the model's ability to detect and correct errors through iterative reasoning.

\begin{figure*}[p]
    \centering
    \includegraphics[width=0.8\textwidth]{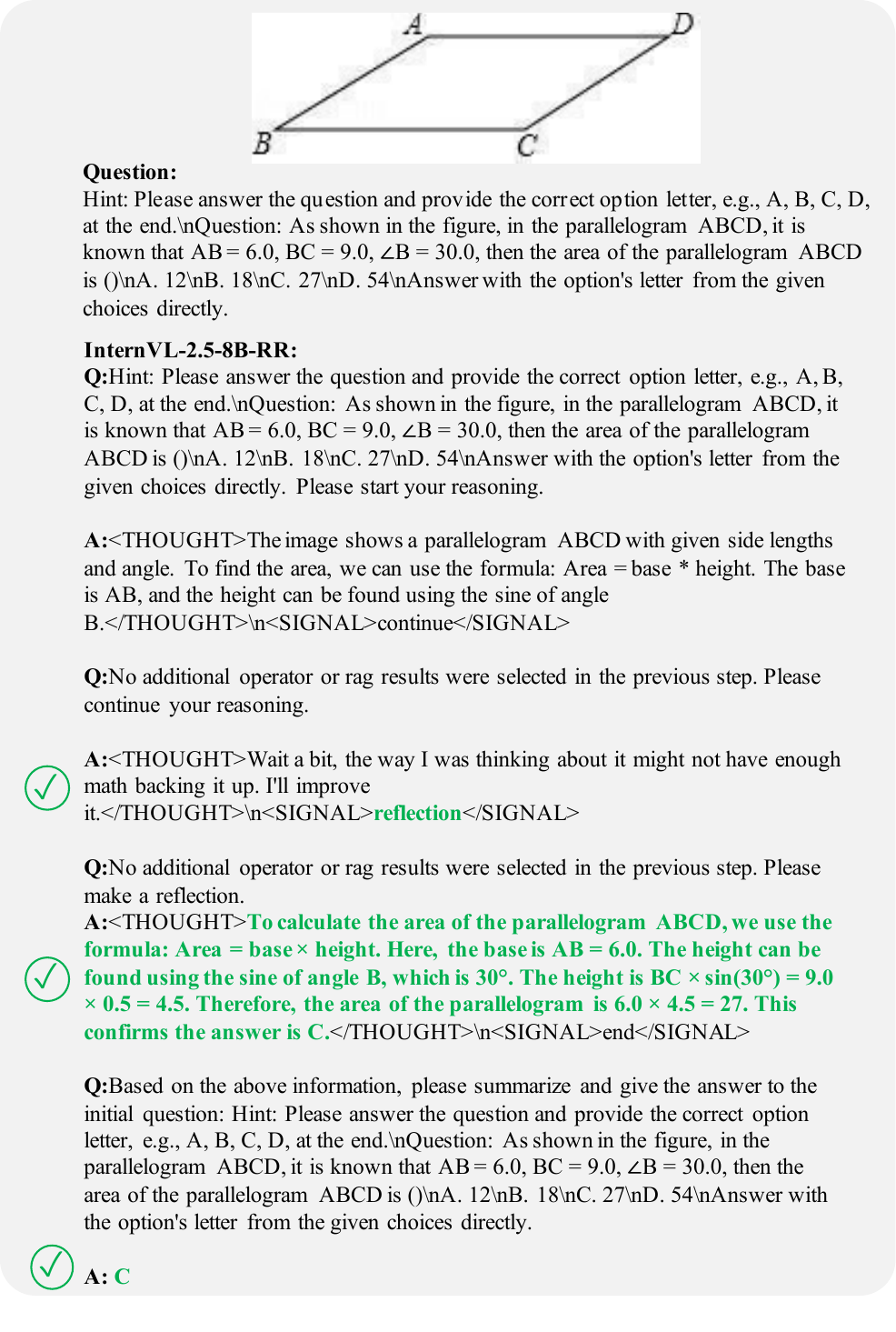}
    \caption{Example of reflection result. The inclusion of the reflection makes the mathematical derivation more complete and accurate.}
    \label{fig:reflection_demo}
\end{figure*}

Figure~\ref{fig:reflection_demo} showcases how reflection improves mathematical reasoning. Initially, the model applies a direct calculation to determine the area of a parallelogram, but realizes its approach lacks mathematical rigor. Through self-reflection, it revises its reasoning and correctly applies the sine function, leading to an accurate computation of the area.

\begin{figure*}[]
    \centering
    \includegraphics[width=0.8\textwidth]{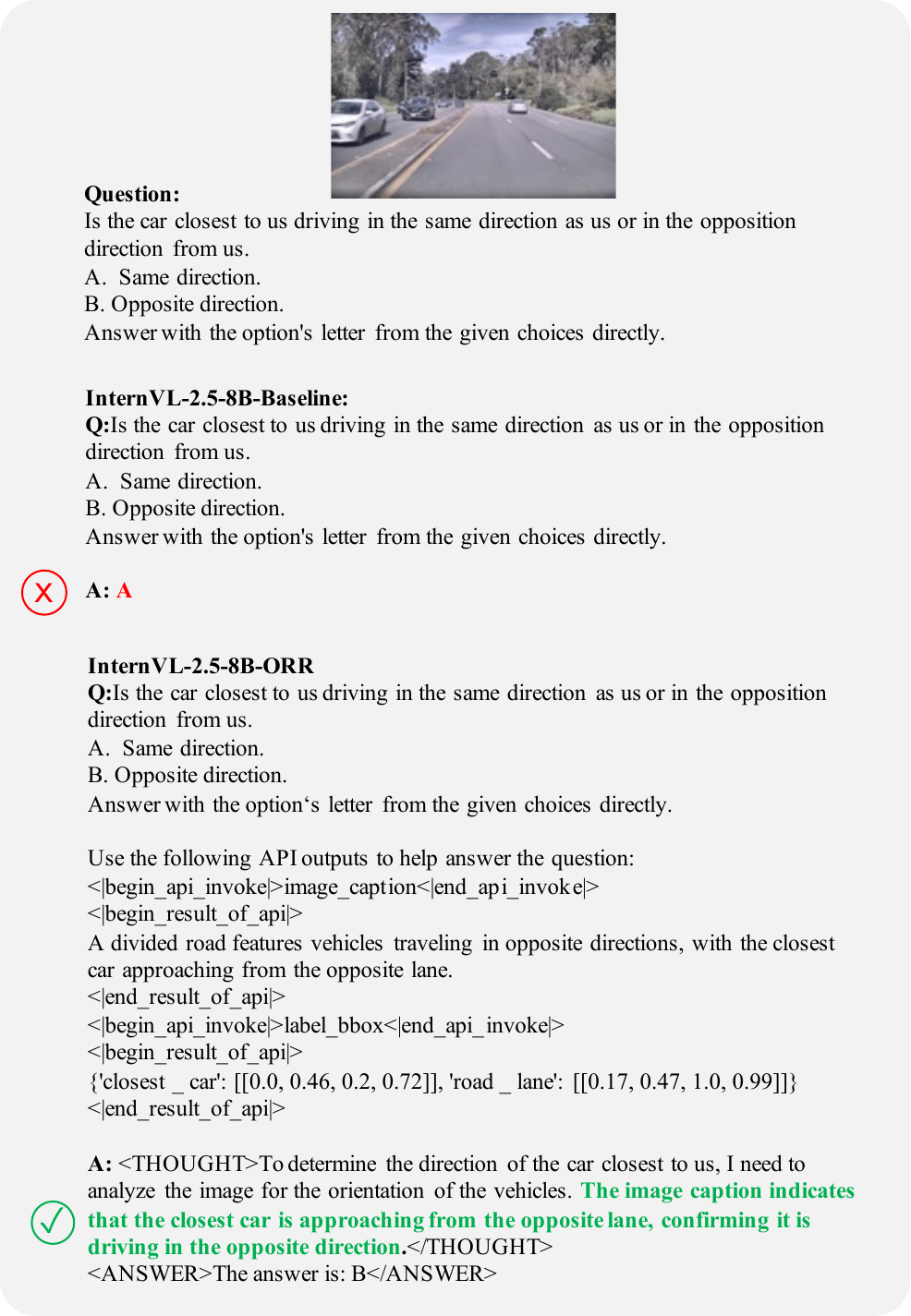}
    \caption{Example of one-turn rationale result. It correctly answers the question with the aid of image caption information.}
    \label{fig:ORR_demo}
\end{figure*}

Figure~\ref{fig:ORR_demo} illustrates how the model integrates external signals such as image caption and object detection to improve decision-making. In a traffic scene, the baseline model incorrectly determines the direction of the closest car. However, by analyzing additional image information, the model correctly identifies that the car is approaching from the opposite lane, leading to the correct conclusion.

These examples underscore the model’s strengths in error correction, reasoning refinement, and effective use of external knowledge for improved decision-making.

\section{Limitations and Future Work}
Despite its strengths, MMAT-1M has certain limitations. The reliance on high-quality rationale training data and a fixed set of tool usage may restrict its generalization ability. Additionally, while the dataset demonstrates strong performance on current benchmarks, further testing is needed to evaluate its adaptability to more MLLMs and more complex scenarios. Nevertheless, MMAT-1M sets a new pipeline for multimodal agent tuning, and its performance on challenging tasks underscores its potential as a foundational visual training dataset for future research.

\clearpage



\end{document}